\def\year{2022}\relax
\documentclass[letterpaper]{article} 
\usepackage{aaai22}  
\usepackage{times}  
\usepackage{helvet}  
\usepackage{courier}  
\usepackage[hyphens]{url}  
\usepackage{graphicx} 
\usepackage{natbib}  
\usepackage{caption} 
\DeclareCaptionStyle{ruled}{labelfont=normalfont,labelsep=colon,strut=off} 
\usepackage{algorithm}

\usepackage{hyperref}       
\usepackage{url}            
\usepackage{booktabs}       
\usepackage{amsfonts}       
\usepackage{nicefrac}       
\usepackage{microtype}      
\usepackage{bm, amsmath}
\usepackage{multirow,array}
\newcommand{\tabincell}[2]{\begin{tabular}{@{}#1@{}}#2\end{tabular}}
\newcommand{\Modelshort}{GRCN }
\usepackage{algorithm}
\usepackage{algpseudocode} 
\usepackage{enumitem}  
\usepackage{pifont}

\newcommand{\cmark}{\ding{51}}%
\newcommand{\xmark}{\ding{55}}%

\usepackage[switch]{lineno}
\usepackage{xcolor}

\usepackage{newfloat}
\usepackage{listings}
\floatstyle{ruled}
\newfloat{listing}{tb}{lst}{}
\floatname{listing}{Listing}
\title{GPN: A Joint Structural Learning Framework for Graph Neural Networks}
\author{
    Qianggang Ding, Deheng Ye, Tingyang Xu, Peilin Zhao\thanks{Corresponding author.}\\
}
\affiliations{
    Tencent AI Lab\\
    Shenzhen, China\\
     \{douglasding, dericye,tingyangxu, masonzhao\}@tencent.com
}

\usepackage{bibentry}

\begin{document}

\maketitle

\begin{abstract}
Graph neural networks (GNNs) have been applied into a variety of graph tasks. Most existing work of GNNs is based on the assumption that the given graph data is optimal, while it is inevitable that there exists missing or incomplete edges in the graph data for training, leading to degraded performance. In this paper, we propose Generative Predictive Network (GPN), a GNN-based joint learning framework that simultaneously learns the graph structure and the downstream task. Specifically, we develop a bilevel optimization framework for this joint learning task, in which the upper optimization (generator) and the lower optimization (predictor) are both instantiated with GNNs. To the best of our knowledge, our method is the first GNN-based bilevel optimization framework for resolving this task. Through extensive experiments, our method outperforms a wide range of baselines using benchmark datasets. 

\end{abstract}

\section{Introduction}

Recently, graph neural networks (GNNs) have attracted much attention due to their strong performance on processing graph structured data. 
GNNs, such as graph convolutional network \cite{kipf2016semi}, graph attention network \cite{velivckovic2017graph}, graph isomorphic network \cite{xu2018powerful}, etc., have been widely used to extract features of nodes and edges in graph data, which can be incorporated to downstream tasks for further calculation. However, it is worth noting that the striking power of GNN heavily depends on the prior structure of graph data. 
As such, GNNs are not robust enough to tackle incomplete and noisy graph structure with missing or excess edges,
resulting in sub-optimal results \cite{yu2019graph}. 
Take the data of citation network \cite{sen2008collective} for an example, a document may not exactly cite all related documents, leading to missing edges. 
Since there exists inevitable incomplete or noisy graph structure in real-world datasets, we believe a direct learning of the specific task is inadequate. 
Hence, in this paper, we are motivated to incorporate the learning of the graph structure and learning of the specific task in one single model, i.e., joint learning. 

\begin{figure}[t]
	\centering
	\includegraphics[width=8.5cm]{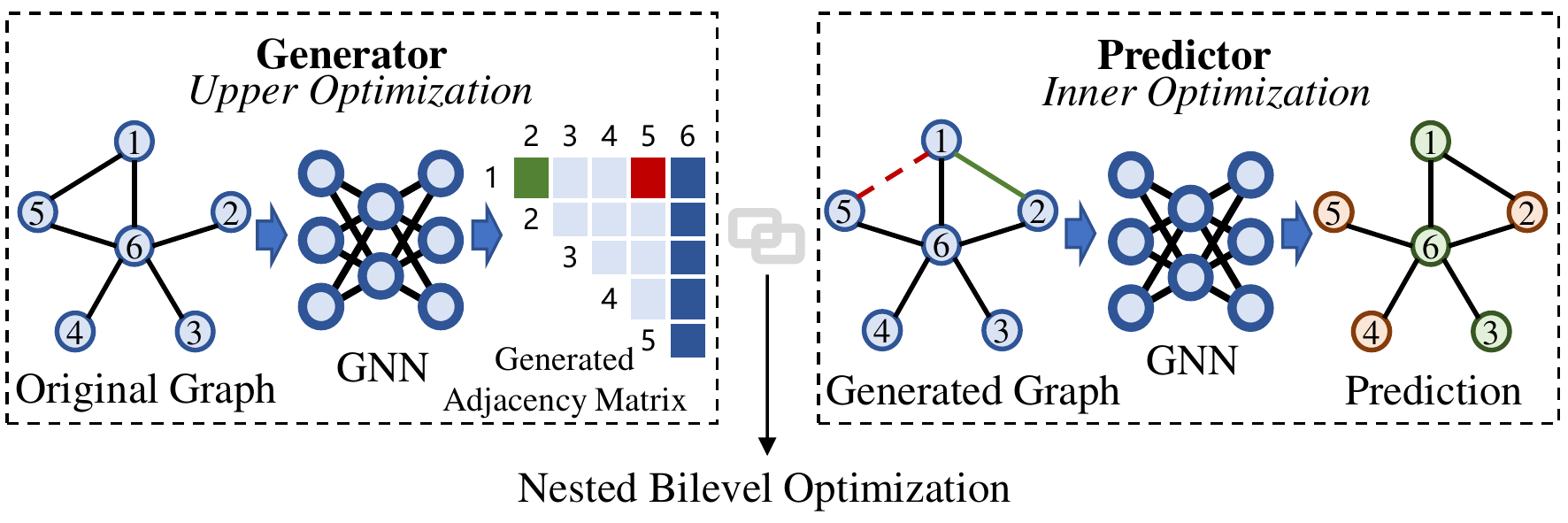}
\vspace{-0.1in}
	\caption{Overview of our proposed GPN: a GNN-based bilevel optimization framework. Both the \textbf{Generator} and the \textbf{Predictor} are instantiated with GNNs. }
	\label{fig:overview}
\vspace{-0.2in}
\end{figure}

However, learning the graph structure implies that we have to modify the adjacency matrix. 
The aforementioned joint learning raises several challenges: 
1) it is naturally a bilevel optimization problem, which is intractable to solve \cite{franceschi2018bilevel}; 
2) parameters of the adjacency matrix are discrete-valued, thus we cannot apply traditional differential optimization methods \cite{ruder2016overview} directly.

To tackle these challenges, the community has proposed several methods, which can be generally divided into two folds: 1) \emph{Bilevel framework}: 
\cite{franceschi2019learning} propose LDS, a novel framework of bilevel optimization, which parameterizes each node pairs and proposes an approximate method to calculate the derivation of them. 
However, the huge amount of parameters make LDS require high memory and time consumption, and the approximate derivation is trivial and inaccurate. Moreover, LDS does not consider the graph specific information while generating the graph structure. 
2) \emph{Non-bilevel framework}: \cite{jiang2019semi} propose GLCN, a novel architecture of GNN integrating both graph learning and graph convolution in a unified GNN architecture. And \cite{yu2019graph} propose GRCN to take an extra GNN to revise the prior graph structure and equips it to the original GNN model. 
However, since these methods solve the bilevel problem by one-stage non-bilevel framework, their architectures are coarse for this problem in some extent. 
Overall, for bilevel-based methods, they neglect the graph specific information while generating the graph structure; and for non-bilevel-based methods, the intrinsic drawbacks of the non-bilevel framework make them not appropriate for the bilevel problem to be solved in this paper. 

To address these concerns, we explore a bilevel-based method with GNN-based generator, which considers the specific graph information while generating the graph structure. Specifically, we decompose the overall joint learning framework into two components: the \textbf{Generator} for the upper optimization and the \textbf{Predictor} for the lower optimization. 
Both of them are instantiated with GNNs and are updated by iterative optimization strategy. 
Instead of directly learning the parameters of the adjacency matrix of graph in LDS, we learn a GNN-based generator to generate the adjacency matrix. 
As a result, our method not only obtains flexibilities from the GNN-based generator in order to handle inductive settings, but also can benefit from the sparsification tricks of GNN to decrease the computation cost to process large graphs. 

In short, the main contributions of this work include:
\begin{itemize}
    \item We propose the first graph-neural-network-based bilevel optimization framework to jointly learn the graph structure and the downstream task. 
    \item We leverage GNN to generate the graph structure in our bilevel framework, and we develop multi-head enhancements to our predictor.
    \item Extensive experiments on semi-supervised node classification tasks show that our method achieves highly competitive results on a range of well-known benchmarks and outperforms state-of-the-art methods. 
\end{itemize}

\section{Related Work}

\paragraph{Graph neural networks} 
Recently, neural networks for graph structured data gain popularity. 
Motivated by the huge success of neural network-based models like CNNs and RNNs, many new generalizations and operations have been developed to handle complex graph data. 
For instance, \cite{kipf2016semi} propose graph convolutional networks (GCN), in which graph convolutional operations are generalized from 2-D convolutional operations of CNNs. Afterwards, a variety of graph operations have been developed rapidly, such as graph attention networks (GAT) \cite{velivckovic2017graph}, Graph-SAGE \cite{hamilton2017inductive}, graph isomorphic networks (GIN) \cite{xu2018powerful}, and Graph Capsule Network~\cite{DBLP:conf/aaai/YangZRYLMH21}. All these graph operations are based on the message-passing mechanism, which takes the weighted average of a node's neighborhood information as the feature of this node. GNNs have been widely used in handling graph structured data. 
However, existing works on GNNs did not fully consider the circumstances of incomplete or noisy graph structure.

\paragraph{Link prediction} Link prediction is an important and tough task in graph theory. In reality, link prediction has numerous real-world applications, such as recommendation systems, knowledge graph analysis, and disease prediction, etc. The aim of link prediction is to predict whether a node is related to another node. The related methods address this problem by measuring the similarities of each pair of nodes. Among them, \cite{wang2007local} propose a statistic method to make a binary decision on the existence of edges. More recently, \cite{liu2017learning} propose probabilistic relational matrix factorization (PRMF), which can automatically learn the dependencies between users in recommendation system. Enhanced by GNN, \cite{zhang2018link} propose a more powerful method outperforms previous methods. However, all these methods only pay attention to complete the graph without considering other tasks such as node classification.

\paragraph{Graph generation} The methods of graph generation can be easily divided into the global methods and the sequential methods. The global methods generate a graph all at once, while the sequential methods generate a graph by outputting nodes or edges one by one. Graph auto-encoders (GAE) is a wide family of global methods of graph generation, which includes Graph-VAE \cite{simonovsky2018graphvae}, Regularized GraphVAE (RGVAE) \cite{ma2018constrained}, and NetGAN \cite{bojchevski2018netgan} etc. The sequential methods \cite{gomez2018automatic}, \cite{kusner2017grammar}, and \cite{dai2018syntax} have attracted more attention in molecular generation through a representation of molecular graph called SMILES. 
Overall, sequential methods linearize a graph as sequences while losing a global perspective on it. Global methods output the whole graph at once while requiring much more computation. 
However, it is unacceptable to simply combine the global or sequential graph generation with the specific downstream task because training the graph generation itself is already very difficult \cite{bojchevski2018netgan}.

\section{Background}

\subsection{Graph Theory}

In general, a graph $G$ can be represented as $G=\{V, E\}$, where $V$ denotes the node set and $E$ denotes the edge set. Typically, given a graph $G$ with $N$ nodes and $M$ edges, we use an adjacency matrix $A \in \{0,1\}^{N\times N}$ to represent edges of graph, where $A_{i,j} = 1$ if there exists an edge from node $v_i$ to node $v_j$, and $A_{i,j} = 0$ otherwise. Depending on the downstream task, each node $v_i$ in $V$ has its own feature, which can be represented by $\bm{x_i} \in \mathbb{R}^{F}$, where $F$ denotes the dimension of node feature. Therefore, we can use a matrix $X \in \mathbb{R}^{N\times F} $ to denote the node features of the whole graph.

\begin{figure*}[t]
	\centering
	\includegraphics[width=15cm]{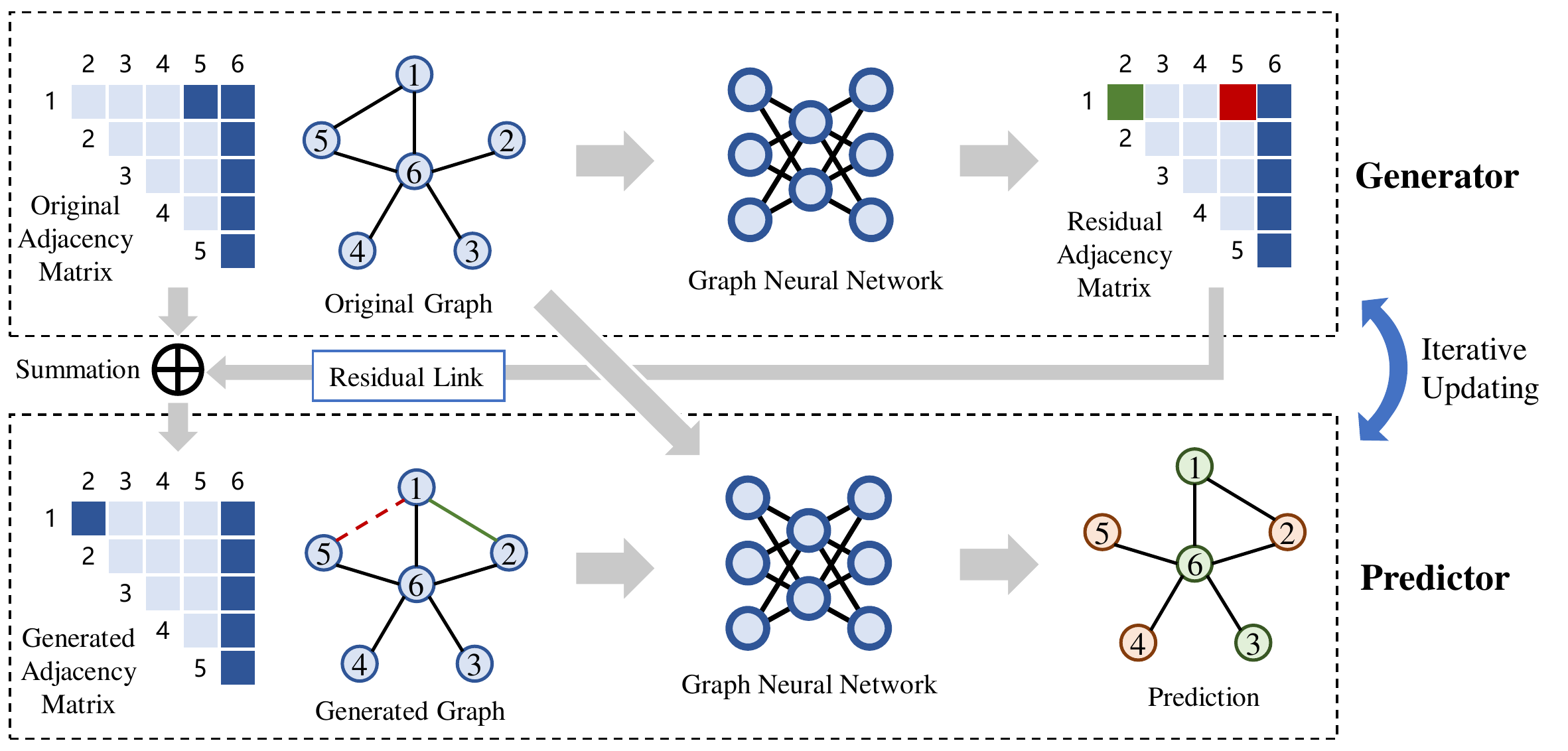}
	\caption{Overall pipeline of GPN.}
	\label{fig:overview}
\vspace{-0.1in}
\end{figure*}

\subsection{Graph Neural Networks}

Recently, graph neural networks (GNNs) achieve a huge success in numerous graph tasks due to their powerful capabilities of extracting features of graphs. For simplicity, we introduce the most common GNN called graph convolutional network (GCN) \cite{kipf2016semi} here. Given a graph $G$ with the adjacency matrix $A$ and the node feature matrix $X$, the propagation procedure of GCN is as the following:
\begin{equation}
    H^{(0)} = X,\  H^{(l+1)} = \sigma(\widetilde{D}^{-\frac{1}{2}}\widetilde{A}\widetilde{D}^{-\frac{1}{2}} H^{(l)} W^{(l)}),
\end{equation}
where $\widetilde{D}$ is a diagonal matrix with $\widetilde{D}_{i,i} = \sum_{j} \widetilde{A}_{i,j}$, $W^{(l)}$ is the trainable parameters of the $l$-th GCN layer, and $\sigma(\cdot)$ is the activation function. After stacking $L$ GCN layers, we can obtain an $L$-layer GCN, which finally outputs the extracted node features $H^{(L)} \in \mathbb{R}^{N \times D}$, where $D$ is the dimension of the extracted node features. Similar to GCN, other variants of GNNs can be represented by the generic paradigm as $H^{(L)} = \textrm{GNN}(X,A)$, which can be utilized for further calculation. 

\subsection{Bilevel Programming}

Distinct from typical optimization problems, the optimization problem of bilevel programming contains a lower level optimization task within the constraint of another upper level optimization task. Formally, given the upper level objective function $F_{\theta}$ parameterized by $\theta$ and the lower level objective function $f_w$ parameterized by $w$, the bilevel optimization problem is given as follows:
\begin{equation}
    \min_{\theta} F_{\theta}(w_\theta, \theta) \ \mathrm{subject}\ \mathrm{to}\ w_\theta \in \mathop{\arg\min}_{w} f_w(w, \theta).
\end{equation}
The nested structure of the bilevel optimization problem requires that a solution to the upper level problem may be feasible only if it is an optimal solution to the lower level problem. This requirement makes bilevel optimization problems very difficult to solve. In the field of machine learning, there are lots of well-known bilevel optimization problems, such as meta-learning \cite{finn2017model}, hyper-parameter tuning \cite{franceschi2018bilevel}, generative adversarial networks (GAN) \cite{goodfellow2014generative}, and network architecture searching \cite{liu2018darts}, etc. In this paper, we introduce a bilevel framework named generative predictive networks (GPN) for graph tasks.

\section{Methodology}

In this work, we model the task of joint learning the graph structure and the downstream task as a bilevel programming problem. The upper level objective is to generate the graph structure by the \textbf{Generator}, and the lower optimization objective is to minimize the training error of the downstream task by the \textbf{Predictor}. Both the predictor and the generator are instantiated with GNNs. To be consistent with the related work \cite{franceschi2019learning} and \cite{yu2019graph}, we specifically focus on the task of semi-supervised node classification in the rest of this paper.

\subsection{Predictor}

The predictor can be instantiated with vanilla GCN-based classifier. Given a graph with $N$ nodes, we consider a $K$-class GNN-based classifier $h_w: \mathcal{X}_N \times \mathcal{H}_N \to \mathcal{C}^N $ as our predictor, where $w$ refers to learnable parameters, $\mathcal{X}_N \subset \mathbb{R}^{N \times F}$ is the space of node features, $\mathcal{H}_N \subset \{0,1\}^{N \times N}$ is the space of the adjacency matrix, and $\mathcal{C} = \{1,2,...,K\}$ is the label space. Given a set of training nodes $V_{train}$, the objective of the predictor is formulated as follows:
\begin{equation}
    \mathop{\arg\min}_{w} f(w, X, A) = \sum_{v\in V_{train}} \mathcal{L}_{h}(h_w(X, A)_v, y_v) + ||w||_2,
\end{equation}
where $h_w(\cdot)_v$ is the predicted label of node $v$, $y_v$ is the ground-truth label of node $v$, and $\mathcal{L}_{h}$ is a point-wise loss function. In this paper, the function $h_w$ can be typically computed as follows:
\begin{equation}
    h_w(X, A) = \textrm{softmax}(\textrm{GNN}_w(X, A)).
\end{equation}
The loss function $\mathcal{L}_h$ can be the cross-entropy loss between $h_w(X,A)_v$ and the ground-truth $y_v$.

\subsection{Generator}

Distinct from LDS \cite{franceschi2019learning} which directly optimizes the adjacency matrix with the proposed approximated hyper-gradient optimization strategy, we use a GNN to generate the adjacency matrix with an iterative updating strategy to optimize the parameters of the GNN. In our method, there is no need to directly optimize the discrete-valued variable which is intractable for common gradient-based optimization methods \cite{franceschi2019learning}. 

To estimate the generalization error of our bilevel model, we aim to minimize the generalization error on another subset of vertices with ground-truth labels, the validation set $V_{val}$. Specifically, for the generator, given a defective initial structure of the graph, i.e. the initial adjacency matrix $A$, it will output a new structure of the graph (i.e. the generated adjacency matrix $\hat{A}$) which is more useful for the predictor. As is the upper-level objective, we can naturally define the objective of the generator with GNN as follows:
\begin{equation}
\begin{aligned}
    &\min_{ \theta} F(w_\theta, \theta, X, A) \\
    &=\sum_{v\in V_{val}} \mathcal{L}_{h}(h_{w_\theta}(X, \hat{A})_v, y_v) + ||w_\theta||_2 + ||\theta||_2, \\
    & \textrm{such that}\ w_\theta = \mathop{\arg\min}_{w} f(w, X, \hat{A})
\end{aligned}
\label{eq:generator}
\end{equation}
where $\hat{A} = A + g_\theta(X, A)$, $g_\theta(X,A) = K(\textrm{GNN}_\theta(X,A))$, $\theta$ are parameters of the generator's GNN and $k(\cdot)$ denotes the kernel function for node embeddings $K: \mathbb{R}^{N \times F} \times \mathbb{R}^{N \times F} \to \mathbb{R}^{N \times N}$. More specifically, given the learned node embeddings $H = \textrm{GNN}_\theta(X,A)$, the kernel function is computed by $K(H)_{i,j} = k(H_i, H_j)$ where $k(\cdot)$ is a similarity function for vectors, such as Euclidean distance, cosine similarity, and dot product. We will explore the choices of $k(\cdot)$ in the later experiments. Note that $g_\theta(X, A)$ represents the residual adjacency matrix and the final generated adjacency matrix is the summation of the initial adjacency matrix and the residual adjacency matrix, which is known as the ``residual link'' operation. 

\subsection{Optimization}

Directly optimizing Eq.~\ref{eq:generator} is intractable for gradient-based methods due to the nested structure and the expensive lower optimization. Therefore, We approximate the gradient of the generator as follows:

\begin{equation}
\begin{aligned}
    & \nabla_{\theta} F(w_\theta^{*}, \theta, X, A) \\
    \approx & \nabla_{\theta} F(w_\theta - \eta \nabla_{w_\theta} f(w_\theta, X, A), \theta, X, A),
\end{aligned}
\label{eq:gradient}
\end{equation}
where $w_\theta^{*}$ denotes the local optimal parameters of the predictor based on the current $\theta$, and $\eta$ is the learning rate for one step of the predictor. Instead of training the predictor to complete convergence, we approximate $w_\theta^{*}$ by training $w_\theta$ only one step at each iteration. This idea has been widely used in meta-learning \cite{finn2017model,DBLP:conf/icml/WeiZH21}, hyper-parameter tuning \cite{franceschi2018bilevel}, generative adversarial networks \cite{goodfellow2014generative}, etc. To increase the robustness, we also introduce the original graph structure to the optimization of the predictor. The predictor is alternately trained on the generated graph structure and the original graph structure. Note that in the inference phase, the predictor only takes the generated graph structure as input. The overall iterative workflow is illustrated in Algorithm~\ref{alg:gpn}. 

\begin{algorithm}
	\caption{Generative Predictive Networks (GPN)} 
	\label{alg:gpn}
	\begin{algorithmic}[1]
	    \State $K(\textrm{GNN}_\theta(X,A)) \gets \bm{0}$ \algorithmiccomment{Initialize the residual adjacency matrix with zero matrix.}
		\For {$epoch=1,2,\ldots$}
		    \State $\theta \gets \theta - \nabla_{\theta} F(w_\theta - \eta \nabla_{w_\theta} f(w_\theta, X, A), \theta, X, A)$ \algorithmiccomment{Update the generator on the validation set. $\eta = 0$ if using FOA approximation.}
		    \State $w_\theta \gets w_\theta - \nabla_{w_\theta} f(w_\theta, X, \hat{A})$ \algorithmiccomment{Update the predictor with the generated graph structure on the training set.}
		    \State $w_\theta \gets w_\theta - \nabla_{w_\theta} f(w_\theta, X, A)$ \algorithmiccomment{Update the predictor with the original graph structure on the training set.}
		\EndFor
	\end{algorithmic} 
\end{algorithm}

\subsection{Approximation}

We further unroll Eq.~\ref{eq:gradient} by applying the chain rule as follows (See Appendix A for more details):
\begin{equation}
    \nabla_{\theta} F(w_\theta', \theta, X, A) - \eta \nabla^2_{w_\theta, \theta} f(w_\theta,X,A) \nabla_{w_\theta'} F(w_\theta', \theta, X, A),
\label{eq:chain}
\end{equation}
where $w_\theta' = w_\theta - \eta \nabla_{w_\theta} f(w_\theta, X, A)$ is the weights for the one-step forward predictor. Observing the second term of Eq.~\ref{eq:chain}, the expensive matrix-vector product will lead to a huge memory and time consumption. Therefore, we propose two approximation methods which can substantially reduce the cost.

\paragraph{Finite Difference Approximation (FDA).} FDA is widely used for solving differential equations by approximating them with difference equations that finite differences approximate the derivatives. Formally, we can transform the second term of Eq.~\ref{eq:chain} to the following equation using FDA (See Appendix B for more details):
\begin{equation}
\begin{aligned}
    & \eta \nabla^2_{w_\theta, \theta} f(w_\theta,X,A) \nabla_{w_\theta'} F(w_\theta', \theta, X, A) \\
    \approx & \eta \frac{\nabla_\theta f(w_\theta^{+},X,A) - \nabla_\theta f(w_\theta^{-},X,A)}{2\epsilon}
\end{aligned}
\end{equation}
where $\epsilon$ is a small scalar, e.g. $\epsilon = 0.01/||\nabla_{w_\theta'} F(w_\theta', \theta, X, A)||$, 
and $w_\theta^{\pm} = w_\theta \pm \epsilon \nabla_{w_\theta'} F(w_\theta', \theta, X, A)$. By this approximation, the expensive product is replaced by the cheap summation, and the complexity is reduced from $O(|w_\theta||\theta|)$ to $O(|w_\theta|+|\theta|)$.

\paragraph{First-Order Approximation (FOA).} In FOA, we directly set $\eta = 0$, then the expensive second term of Eq.\ref{eq:chain} will disappear accordingly. By this approximation, we assume the current $w_\theta$ as the local optimal $w_\theta^{*}$, therefore FOA has less complexity than FDA. 

\begin{figure}[t]
	\centering
	\includegraphics[width=8cm]{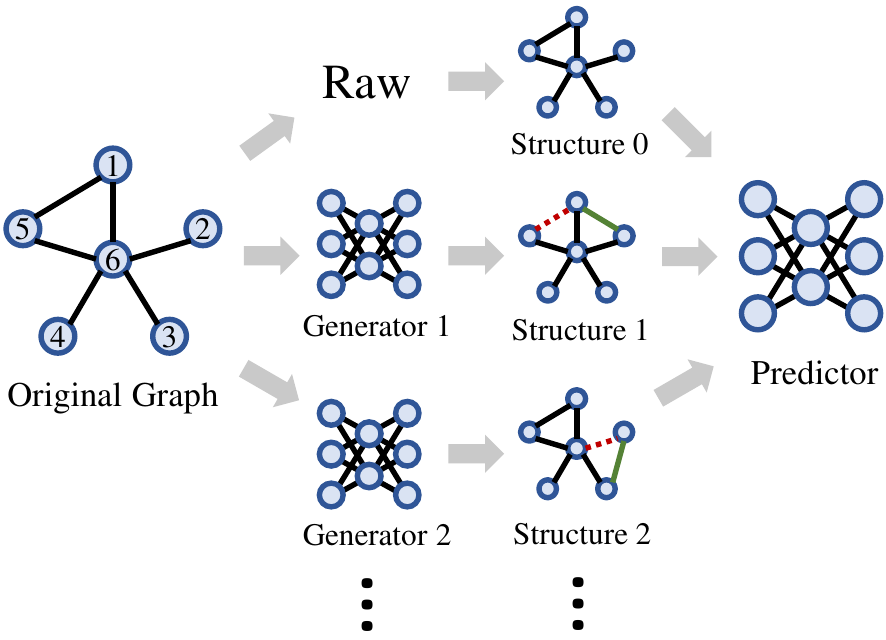}
	\caption{Multi-head enhancement.}
	\label{fig:multi}
	\vspace{-0.2in}
\end{figure}

\subsection{Multi-head Enhancement}
\label{sec:multi}

To increase the robustness of GPN, we propose multi-head GPN enhanced by multi-branch generator. Instead of the original single-head generator which generates only one graph structure at one iteration, the multi-head generator is able to generate multiple graph structures simultaneously. The rationale behind this idea is that there may exist multiple optimal structures for a given graph. For the proposed multi-branch architecture, please refers to Fig.~\ref{fig:multi}. An $N$-branch generator will generate $N$ structures and the predictor takes all these structures as inputs to train the parameters. By generating multiple structures at the same time, we can produce more diverse inputs for the predictor, which further improve the predictor's robustness. Our empirical results show an extra gain achieved by this enhancement. 

\begin{figure*}[t]
	\centering
	\includegraphics[width=\textwidth]{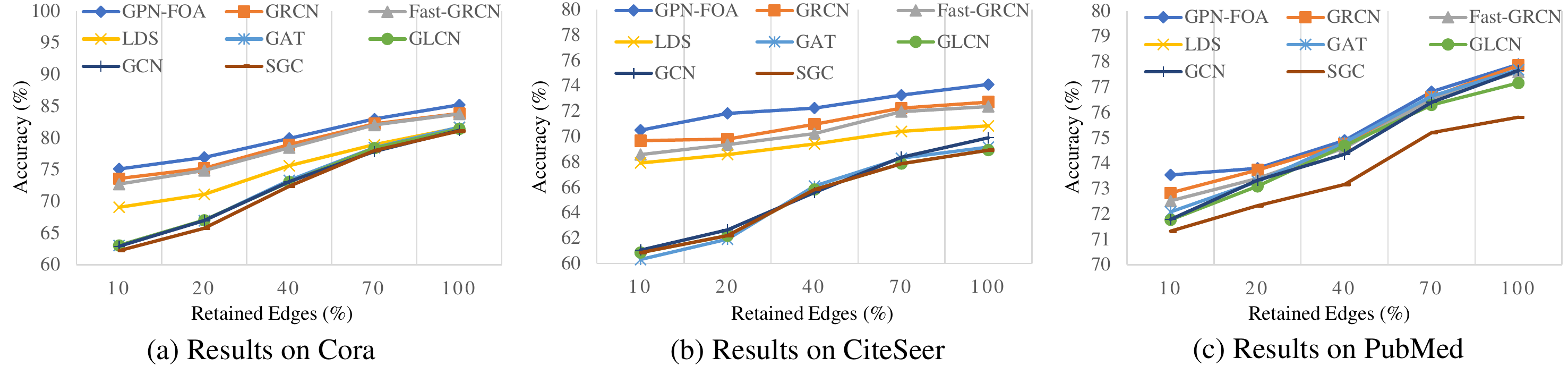}
	\caption{Results on different ratios of incomplete edges.}
	\label{fig:incomplete}
\vspace{-0.1in}
\end{figure*}

\section{Experiments}

Semi-supervised node classification is a commonly used task to demonstrate the usages of graph models. We evaluate our methods (GPN-FOA, GPN-FDA) on a set of semi-supervised node classification tasks. Then we conduct experiments for exploring the choices of different kernel functions and the choices of different GNNs for the generator. Finally, we perform ablation studies. 

\begin{table}[]
\centering
\caption{Dataset statistics.}
\vspace{-0.1in}
\begin{tabular}{@{}ccccc@{}}
\toprule
Datasets  & Nodes & Edges & Features & Classes \\ \midrule
Cora      & 2708  & 5429  & 1433     & 7       \\
Citeseer  & 3327  & 4732  & 3703     & 6       \\
Pubmed    & 19717 & 44338 & 500      & 3       \\
Cora-Full & 19793 & 65311 & 8710     & 70      \\
Amazon-Computers & 13381 & 245778 & 767   & 10  \\
Coauthor-CS   & 18333 & 81894 & 6805   & 15  \\ \bottomrule
\end{tabular}
\label{tab:datasets}
\end{table}

\begin{table}[]
\centering
\caption{\label{tab:tab-1} Results on the fixed setting (\%): Mean test classification accuracy and standard deviation in percent averaged for over 5 runs. The best results are marked in \textbf{bold}. N/A denotes the method cannot process the dataset with full-batch inputs due to the memory limitations of GPU. (Same below)}
\vspace{-0.1in}
\begin{tabular}{lccc}
\toprule
\multicolumn{1}{l}{\multirow{2}{*}{Methods}} & \multicolumn{3}{c}{Datasets} \\ \cmidrule(l){2-4} 
\multicolumn{1}{c}{}     & Cora   & CiteSeer  & PubMed  \\ \midrule
GCN  &  $81.4\pm0.5$    &   $70.9\pm0.5$       &  $79.0\pm0.3$      \\
SGC  &   $81.0\pm0.0$   &  $71.9\pm0.1$        & $78.9\pm0.0$       \\
GAT  &   $83.2\pm0.7$   &   $72.6\pm 0.6$       &   $78.8\pm0.3$     \\ 
Graph U-Net  &   $83.7\pm1.5$   &   $73.1\pm 1.8$       &   $77.7\pm0.9$     \\
GraphMix  &   $83.9\pm0.6$   &   $74.5\pm 0.6$       &   \bm{$80.9\pm0.5$}     \\
\midrule
LDS  &   $84.0\pm0.4$   &   $74.8\pm0.5$       &   N/A     \\
GLCN &  $81.8\pm0.6$    &    $70.8 \pm 0.5$      &    $78.8\pm0.4$     \\ 
Fast-\Modelshort & $83.6\pm0.4$    &  $72.9\pm0.6$     &  $79.0\pm0.2$    \\
\Modelshort &   $84.2 \pm 0.4$  &    $73.6 \pm 0.5$      &  $79.0\pm0.2$     \\ \midrule
GPN-FOA &  \bm{$85.9\pm0.3$}    &    \bm{$75.2 \pm 0.2$}      &    $79.2\pm0.1$     \\ 
GPN-FDA &  $85.4\pm0.4$    &    $75.0 \pm 0.3$      &    $79.1\pm0.1$     \\ 
\bottomrule
\end{tabular}
\vspace{-0.2in}
\end{table}

\begin{table}[]
\centering
\caption{\label{tab:tab-2}Results on the random setting (\%).}
\vspace{-0.1in}
\begin{tabular}{lccc}
\toprule
\multicolumn{1}{l}{\multirow{2}{*}{Methods}} & \multicolumn{3}{c}{Datasets} \\ \cmidrule(l){2-4} 
\multicolumn{1}{c}{}     & Cora   & CiteSeer  & PubMed  \\ \midrule
GCN  &  $81.2\pm1.9$    &   $69.8\pm1.9$       &  $77.7\pm2.9$      \\
SGC  &   $81.0\pm1.7$   &  $68.9\pm2.0$        & $75.8\pm3.0$       \\
GAT  &   $81.7\pm1.9$   &   $68.8\pm 1.8$       &   $77.7\pm3.2$     \\
\midrule
LDS  &   $81.6\pm1.0$   &   $71.0\pm0.9$       &   N/A     \\
GLCN &  $81.4\pm1.9$    &    $69.8 \pm 1.8$      &    $77.2\pm3.2$     \\ 
Fast-\Modelshort &   $83.8 \pm 1.6$  &    $72.3 \pm 1.4$      &    $77.6 \pm 3.2$   \\
\Modelshort &   $83.7 \pm 1.7$  &    $72.6 \pm 1.3$      &    $77.9 \pm 3.2$   \\ \midrule
GPN-FOA &   \bm{$85.5 \pm 1.4$}  &    \bm{$74.5 \pm 1.5$}      &  \bm{$78.2 \pm 3.0$}   \\ 
GPN-FDA &   $84.3 \pm 1.5$  &    $72.8 \pm 1.6$      &  $77.8 \pm 2.6$   \\
\toprule
\multicolumn{1}{l}{\multirow{2}{*}{Methods}} & \multicolumn{3}{c}{Datasets} \\ \cmidrule(l){2-4} 
\multicolumn{1}{c}{}     & Cora-Full   & \tabincell{c}{Amazon\\Computers}  & \tabincell{c}{Coauthor\\CS}  \\ \midrule
GCN  &  $60.3\pm0.7$    &   $81.9\pm1.7$       &   $91.3\pm0.3$      \\
SGC  &  $59.1\pm 0.7$    &   $81.8 \pm 2.3$       &     $91.3\pm0.2$    \\
GAT  &  $59.9\pm 0.6$    &    $81.8\pm2.0$      &   $89.5\pm 0.5$     \\ \midrule
LDS  &   N/A   &    N/A      &  N/A      \\
GLCN &  $59.1\pm 0.7$    &  $80.4\pm1.9$        &  $90.1\pm 0.5$      \\ 
Fast-\Modelshort &   $60.2\pm0.5$   &    $83.5\pm1.6$    & $91.2\pm0.4$ \\
\Modelshort &   $60.3\pm0.4$   &    $83.7\pm1.8$      & $91.3\pm0.3$ \\ \midrule
GPN-FOA &   \bm{$61.3 \pm 0.5$}  &    \bm{$84.6 \pm 1.8$}      &  \bm{$91.8 \pm 0.5$}   \\ 
GPN-FDA &   $60.8 \pm 0.5$  &    $84.0 \pm 1.6$      &  $91.4 \pm 0.5$   \\
\bottomrule
\vspace{-0.2in}
\end{tabular}
\end{table}

\subsection{Datasets}

We use the following graph datasets for the semi-supervised node classification task specifically: Cora, Citeseer \cite{sen2008collective}, and PubMed \cite{namata2012query}, which are three common benchmarks for node classification. We follow the settings mentioned in \cite{yu2019graph}, which consists of the fixed setting and the random setting. For the fixed setting, we split the train/valid/test sets followed the rule mentioned in previous work \cite{yang2016revisiting}. For the random setting, we randomly split the train/valid/test sets with the same number of labels with the fixed settings. To further evaluate the performance on large graphs, we evaluate our methods with baselines on several larger datasets: Cora-Full, Amazon-Computers, and Coauthor-CS. For these datasets, we follow the settings mentioned in \cite{yu2019graph} and \cite{shchur2018pitfalls}, which takes 20 labels of each classes for training, 30 for validating, and the rest for testing. Classes with less than 50 labels are deleted from the datasets. 
For the overall description of datasets, please refer to Table~\ref{tab:datasets}. Furthermore, we also evaluate our methods on these datasets with different ratios of incomplete edges, which are created by randomly dropping some edges. These experiments will further demonstrate the superiority of our methods.

\subsection{Implementation Details}

We implement our GPN family with PyTorch library \cite{paszke2019pytorch} on a single Nvidia Tesla V100 GPU (32GB on-board memory). For Cora, PubMed, Amazon-Computers, and Coauthor-CS datasets, we use Adam optimizer with an initial learning rate of $0.005$. For CiteSeer, the initial learning rate is set to $0.05$. For all datasets, the weight decay of the optimizer is set to $0.005$. We pre-train parameters of both the generator and the predictor with a non-bilevel version of GPN, where the upper optimization and the lower optimization share a coordinate descent strategy instead of the iterative descent strategy. The epochs of the pre-training phase and the main training phase are both set to $300$.

\begin{table}[]
\centering
\caption{Results of the inductive setting (\%). N/A denotes this method cannot apply to this setting. We train each model on Cora dataset with partial nodes ranging from $20\%$ to $80\%$, then test the model on $100\%$ nodes.}
\label{tab:nodes}
\begin{tabular}{@{}lcccc|c@{}}
\toprule
\multirow{2}{*}{Methods} & \multicolumn{5}{c}{Ratio (\%)} \\ \cmidrule(l){2-6} 
                         & 20   & 40  & 60  & 80  & 100  \\ \midrule
GCN                      &$49.6$&$77.5$&$79.1$&$80.1$&$81.4$\\
GRCN                     &$56.2$&$82.6$&$83.0$&$83.1$&$84.2$     \\
LDS                      & N/A  & N/A & N/A & N/A & $84.0$ \\
GPN-FOA                  &\bm{$66.5$}&\bm{$83.8$}&\bm{$84.2$}&\bm{$84.9$}&\bm{$85.9$}      \\ \bottomrule
\end{tabular}
\end{table}

\subsection{Baselines}

We compare our methods with the following competitive baselines: 

\begin{itemize}
    \item GCN \cite{kipf2016semi}:  a common GNN used as the benchmark for node classification tasks.
    \item GAT \cite{velivckovic2017graph}:  an advanced GNN enhanced with the attention mechanism, the state-of-the-art GNN model on most graph datasets.
    \item SGC \cite{wu2019simplifying}:  a variant of GNNs without non-linear layers, which has much less complexity than traditional GNNs.
    \item Graph U-Net \cite{gao2019graph}: a novel GNN architecture with the proposed attention-based pooling layers.
    \item GraphMix \cite{verma2019graphmix}: a regularized training scheme for vanilla GNNs, leveraging the advances of classical DNNs.
    \item LDS \cite{franceschi2019learning}:  a probabilistic generative model under the bilevel framework for graph structure learning.
    \item GLCN \cite{jiang2019semi}: a GNN-based model integrating both graph learning and graph convolution in a unified GNN architecture.
    \item GRCN \cite{yu2019graph}: a GNN-based model with an extra GNN-based reviser to revise the prior graph structure.   
\end{itemize}

\subsection{Main Results}

We compare the performance of our methods (GPN-FOA, GPN-FDA) with other baselines on Cora, CiteSeer, PubMed, Amazon-Computers, and Coauthors-CS datasets. Among them, GPN-FOA denotes the method with first-order approximation, GPN-FDA denotes the method with finite difference approximation. In all experiments of GPN, the similarity function $k(\cdot)$ is the dot-product similarity. To make a fair comparison to other baselines (GRCN, LDS), the generator and the predictor in our methods are instantiated with GCNs.

The results are summarized in Table~\ref{tab:tab-1}, from which we can observe that our GPN-based methods outperform other baselines on Cora and CiteSeer datasets, and achieve competitive performance on PubMed datasets. Specifically, compared with traditional GNNs, GPN-based methods achieve ~1.2\% gains over GAT, one of the state-of-the-art GNNs, on Cora datasets. Meanwhile, compared with LDS, our better performance shows the effectiveness of our GNN-based generator over parameterized probabilistic sampling. Compared with GRCN, which is the strongest baseline, our method also significantly outperform it on Cora and CiteSeer datasets, which shows the advantage of our bilevel framework.

We further conduct some experiments with different ratios of edge incompleteness. The comparison with other baselines can be seen in Fig.~\ref{fig:incomplete}. We can observe that our methods consistently outperform other baselines, which exhibits the robustness of GPN.

\subsection{Inductive Experiments}

In this section, we compare the transfer capability of GPN with other baselines. Specifically, we train each model on the graph with only partial nodes and then test the model on the entire complete graph. The results are shown in Table~\ref{tab:nodes}. From the results, we can observe that GPN outperforms all baselines consistently. It is worth nothing that LDS can only be used for transductive setting, therefore it cannot handle new nodes during the testing. Moreover, it is interesting to note that GPN could obtain better results over GCN using only $40\%$ nodes, and obtain competitive results with GRCN using only $60\%$ nodes. The results demonstrate the effectiveness of the generator of GPN, since the only difference between GPN and GCN is that GPN uses the graph generated by the generator as the input graph for its predictor, while GCN takes the prior graph. 
Overall, these results demonstrate that GPN could be applied to the cases with the need of adding nodes online, so that it is more attractive than LDS.

\subsection{Multi-head GPN}

As mentioned, the multi-head enhancement aims to generate multiple graph structures simultaneously in order to improve the robustness of the predictor. We conduct several experiments with different number of branches of the generator. Note that we select the branch of generator with the best performance on the validation set in the testing phase. From Table~\ref{tab:multi}, we can observe that the increase of heads could further improve the performance of GPN, which exhibits that the diversity of generated graph contributes to the effectiveness of the predictor. 

\begin{table}[]
\centering
\caption{\label{tab:multi} Results of multi-head enhancement (\%).}
\begin{tabular}{@{}lccc@{}}
\toprule
\multicolumn{1}{l}{\multirow{2}{*}{\# Heads}} & \multicolumn{3}{c}{Datasets} \\ \cmidrule(l){2-4} 
\multicolumn{1}{c}{}     & Cora   & CiteSeer  & PubMed  \\ \midrule
1 (default)              & $85.9\pm0.3$ & $75.2\pm0.2$ & $79.2\pm0.1$ \\
2                        & $86.0\pm0.2$ & $75.2\pm0.2$ & $79.1\pm0.1$  \\
4                        & \bm{$86.2\pm0.2$} & $75.4\pm0.2$ & $79.2\pm0.1$  \\ 
8                        & \bm{$86.2\pm0.3$} & \bm{$75.8\pm0.2$} & \bm{$79.3\pm0.1$}  \\ \bottomrule
\end{tabular}
\end{table}

\begin{table}[]
\centering
\caption{\label{tab:similarity} Results on choices of the similarity function (\%). N/A denotes the form of similarity function cannot process the dataset with full-batch inputs due to the memory limitations of GPU.}
\begin{tabular}{@{}llll@{}}
\toprule
\multicolumn{1}{l}{\multirow{2}{*}{Methods}} & \multicolumn{3}{c}{Datasets} \\ \cmidrule(l){2-4} 
\multicolumn{1}{c}{}     & Cora   & CiteSeer  & PubMed  \\ \midrule
Cosine            & $61.0\pm0.7$ & $54.2\pm0.6$ & N/A \\
Euclidean         & $34.5\pm0.6$ & N/A & N/A  \\
Dot-product       & \bm{$85.9\pm0.3$} & \bm{$75.2\pm0.2$} & \bm{$79.2\pm0.1$}  \\ \bottomrule
\end{tabular}
\end{table}

\subsection{Ablation Study}

We conduct several ablation studies: 1) the effectiveness of choices for the backbone GNN of the generator, 2) the form of the similarity function $k(\cdot)$, and 3) the effectiveness of the bilevel framework and non-bilevel pretraining. 

\paragraph{Similarity function.} We explore three forms of the similarity function, including dot-product similarity, cosine similarity, and Euclidean similarity. The comparison results on three datasets are shown in Table~\ref{tab:similarity}, which exhibits the superiority of the dot-product similarity on both the memory complexity and the accuracy performance.

\paragraph{GNN of the generator.} We instantiate the generator with other GNNs like GAT, GraphSage, GIN, etc. We attempt to explore whether these empirically powerful GNNs can benefit the generator to produce better graph structure. 
The results are in Table~\ref{tab:generator}. 
We can see that the generators instantiated with GAT, GraphSage, and GIN have consistently worse performance than the one instantiated with GCN, though in general tasks they exhibit more considerable performance. It is also interesting to note that the more complex the GNN is, the easier the over-fitting phenomenon happens. 
Among these experiments, the least complex network, GCN, avoids the over-fitting issue to the largest extent.

\paragraph{Bilevel framework and non-bilevel pretraining.} To better understand the effectiveness of the bilevel optimization, we design a non-bilevel version of GPN, where the predictor and the generator are not trained iteratively but trained simultaneously in one updating phase. This means the two GNNs are updated using coordinate descent strategy on the training set. Here we conduct experiments to explore the effectiveness of the bilevel framework (bilevel) and the non-bilevel pretraining (pretrain). The results in Table~\ref{tab:bilevel} exhibit the superiority of both the bilevel framework and the pretraining.

\begin{table}[]
\centering
\caption{\label{tab:generator} Results on choices of the backbone GNN of the generator (\%).}
\begin{tabular}{@{}llll@{}}
\toprule
\multicolumn{1}{l}{\multirow{2}{*}{GNNs}} & \multicolumn{3}{c}{Datasets} \\ \cmidrule(l){2-4} 
\multicolumn{1}{c}{}                      & Cora   & CiteSeer  & PubMed  \\ \midrule
GCN                                       & \bm{$85.9\pm0.3$} & \bm{$75.2 \pm 0.2$} & \bm{$79.2\pm0.1$} \\
GAT                                       & $84.2\pm0.3$ & $72.2\pm0.4$ & $79.0\pm0.2$  \\
GraphSage                                 & $83.6\pm0.5$ & $73.4\pm0.4$ & \bm{$79.2\pm0.2$}  \\
GIN                                       & $83.6\pm0.5$ & $71.2\pm0.8$ & $79.0\pm0.2$ \\ \bottomrule
\end{tabular}
\end{table}

\begin{table}[]
\centering
\small
\caption{\label{tab:bilevel} Results of the effectiveness of bilevel framework (\%): \textbf{bilevel} means the bilevel framework and \textbf{pretrain} means the pretraining on the non-bilevel version of GPN.}
\begin{tabular}{@{}cc|ccc@{}}
\toprule
\multicolumn{2}{c|}{Settings} & \multicolumn{3}{c}{Datasets} \\ \midrule
bilevel      & pretrain      & Cora   & CiteSeer  & PubMed  \\ \midrule
\cmark     &    \cmark       & \bm{$85.9\pm0.3$} & \bm{$75.2\pm0.2$} & \bm{$79.2\pm0.1$} \\
\cmark     &    \xmark       & $85.5\pm0.3$ & $74.9\pm0.4$ & $79.0\pm0.2$ \\
\xmark     &    \xmark       & $84.9\pm0.5$ & $74.2\pm0.8$ & $79.0\pm0.2$ \\ \bottomrule
\end{tabular}
\label{tab:my-table}
\end{table}

\section{Conclusion and Future Work}

In this paper, we propose Generative Predictive Network (GPN), a bilevel and GNN-based framework for joint learning the graph structure and the downstream task. 
Our GNN-based generator can utilize the graph specific information to generate a new graph structure, and our proposed nested bilevel framework is appropriate and effective for the bilevel setting. 
We have also proposed a multi-head version of GPN, in which the multi-branch generators can generate graph structures simultaneously, so as to further increase the robustness and performance of the predictor. 
Empirical results on several well-known benchmarks show that our method achieves significant gains over state-of-the-art methods. In the future, we will focus on the following aspects: 1) the scalability for large graphs, 2) the capability of handling dynamic graphs, and 3) applying the method to real-world applications such as recommendation systems. 

\newpage
\bibliography{GPN}

\end{document}